\documentclass{article}

\usepackage[creativeai, final]{neurips_2026}
\usepackage[utf8]{inputenc}
\usepackage[T1]{fontenc}
\usepackage{hyperref}
\usepackage{url}
\usepackage{booktabs}
\usepackage{amsfonts}
\usepackage{amsmath}
\usepackage{nicefrac}
\usepackage{microtype}
\usepackage{xcolor}
\usepackage{graphicx}
\graphicspath{{figures/}}
\usepackage{algorithm}
\usepackage{algpseudocode}
\usepackage[normalem]{ulem}

\newcommand{\motif}[1]{\texttt{(#1)}}
\newcommand{\method}{MotiGen}

\title{Getting Motif-ated: Controllable AI Compositions from Injected Motif Prompts}

\author{%
  Chao Peter Yang \\
  Yale University\\
  New Haven, CT\\
  \texttt{chaopeter.yang@yale.edu} \\
  \And
    Cynthia Rudin \\
  Duke University\\
  Durham, NC\\
  \And
  Yue Jiang\\
  Duke University\\
  Durham, NC\\
  \AND
  Simon Mak\\
  Duke University \\
  Durham, NC
  \And
  Stephen Ni-Hahn \\
  Chinese University of Hong Kong, Shenzhen \\
  Shenzhen, Guangdong \\
}

\begin{document}

\maketitle

\begin{abstract}
Deep learning has transformed symbolic music generation by borrowing the training paradigms of large language models, with systems such as NotaGen now producing complete, stylistically convincing classical scores from a short prompt. These systems could become powerful creative partners, helping musicians generate endless possibilities. However, current systems expose almost no control handles on the music itself. In principle, control handles could be built into a foundation model trained from scratch, but this is rarely practical without massive amounts of quality annotated data and compute resources. We therefore present \method{}, a recipe for retrofitting pretrained symbolic music models to use new instruction prompts. \method{} injects a musical motif as a structured prompt line, reinforces it with a scalar attention bias toward the motif tokens, and learns the association with a two-phase curriculum. First, it learns from focused excerpts cropped around motif occurrences in the training data, then full scores including the motifs. Our experiments show that our model composes with the prompted motif in over 92.3\% of generated pieces. Generated pieces using a variety of motifs are included in our sample site: \url{https://motigen-site.github.io/}.
\end{abstract}

%To exemplify our methodology, we instantiate it with the most natural unit of musical thought of the \emph{motif}, a short melodic pattern specified as a transposition-invariant sequence of notes or intervals.

%The realistic path to controllability is therefore to graft new control mechanisms onto an already-pretrained model.

%Control is limited to coarse metadata such as period, composer, and instrumentation, so a musician who has a melodic idea in mind has no way to make the model develop it, and naïvely pasting the idea into the prompt of a pretrained model forces the phrase to open the piece, greatly constraining how the idea can be used.

%State-of-the-art systems like NotaGen do not release their training corpora, and reproducing their scale of data and compute is out of reach for most researchers.

\section{Introduction}
\label{sec:intro}

Motifs are the atoms of musical composition. The four notes that open Beethoven's Fifth Symphony are among the most recognizable sounds in Western music, and nearly every bar of the movement grows out of them. Schoenberg called the motif the germ from which coherent musical form develops \citep{schoenberg1967fundamentals}. When composers have a musical idea, it usually arrives in exactly this form, where a handful of notes with a characteristic contour are developed into full-fledged pieces. Meanwhile, autoregressive language models trained on large score corpora \citep{huang2019music, payne2019musenet, wu2023tunesformer, qu2024mupt} now generate long, well-formed pieces. NotaGen in particular \mbox{\citep{wang2025notagen}} showed that the large language model (LLM) paradigm of pretraining, finetuning, and reinforcement learning carries over to musical scores in ABC notation, a textual representation of written music.%, with generated pieces rating competitively with human-composed classical music.

% \cmtS{we might want to be a bit more specific on limitations of existing approaches (avoid jargon, e.g., ``enharmonic spelling''). what are fundamental bottlenecks that prevent musicians from effectively using current AI tools for music composition? this then highlights the importance of our contributions.} 

However, few models give composers the option to control generation at the motif level, which is the granularity that would allow composers to develop their musical ideas into full compositions. Existing control sits at two extremes, either having coarse controls such as style or genre, or needing fully formed musical material as input. Notagen \citep{wang2025notagen}, for example, limits the users to coarse inputs such as composer style, period, and instrumentation, offering no control over the development of musical material. Alternatively, the user might be expected to provide a concrete, full-fledged melody\citep{yuan2024chatmusicianunderstandinggeneratingmusic} to harmonize or a bar-level description sequence \citep{vonrutte2023figaro} that requires fully-composed material.

A motif, in its raw form, is an abstraction that is transposition invariant and may not even have defined intervals, so conditioning on a precise sequence of notes would limit musical generation. In trying to na\"ively prompt existing models, we find that pretrained models ignore prompt lines they were not trained for, so models must be tuned such that additional instructions are followed without degrading the backbone's capabilities in generating music. Furthermore, judging whether a generated piece incorporates the motif appropriately requires specialized evaluation methods.

We close this gap with \method{}. \method{} enables composers to build melodies and full pieces from a single abstract motif, represented as a transposition-invariant interval-class sequence of steps, skips, and leaps together with their contours. The motif is injected as a structured \texttt{\%motif:} prompt line in the score header. A scalar \emph{attention bias} at the motif line's key positions keeps the motif salient throughout decoding. A two-phase \emph{curriculum} teaches the model how to incorporate the motif. Phase~1 fine-tunes on focused excerpts cropped from source pieces around located motif occurrences; phase~2 fine-tunes on full scores annotated with their extracted motifs. The result is a model that allows musicians to develop ideas naturally while retaining significant control over the piece's content. %four notes. %\footnote{While our motif identification algorithm supports various lengths, the authors has found that 4 note motifs tend to be the most abundant and easy to use}

%(e.g., \motif{0,3,-1,-1} would be a note, a leap up, then two steps down)

%, where the motif--content association is strong and local

Our novel contributions include: (1) a motif-conditioning method for symbolic music LLMs combining \texttt{\%motif:} prompt lines with a train- and inference-time attention bias, (2) a focused-crop curriculum that teaches motif grounding from phrases containing each motif and (3) a transposition-invariant, enharmonic-aware motif containment evaluator shared between training annotation and evaluation.

%, including a dose--response characterization of the bias strength

%(step/skip/leap classes, repeat collapse)

% adoption versus unconditioned base rates, ablations of bias and curriculum, the dependence of adoption on training-set motif frequency, and a Gradio co-composition interface. A real-data scaling study appears in Appendix~\ref{app:scaling}.

\begin{figure}[t]
  \centering
  \includegraphics[width=\linewidth]{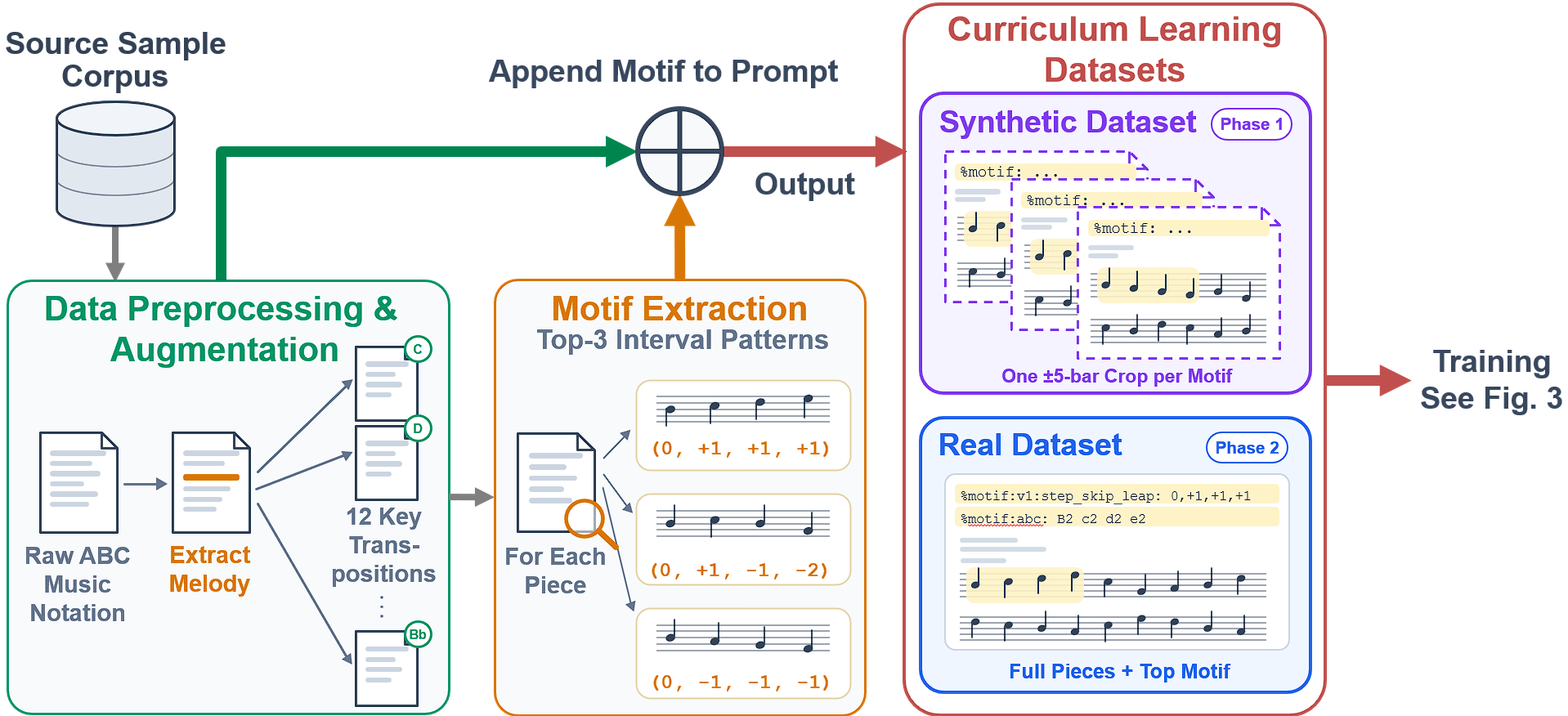}
  \caption{\method{} data pipeline: each piece is reduced to its melody line,
  augmented across keys, and mined for its recurring interval patterns; the
  located occurrences yield the phase-1 focused set of motif-centered crops
  and the phase-2 of full pieces with \texttt{\%motif:} annotations.} 
  \label{fig:overview}
\end{figure}

\section{Related Work}
\label{sec:related}

\subsection*{Generative Symbolic Music Models}
Autoregressive Transformers have long been applied to event-token and text-based score representations, from Music Transformer \citep{huang2019music} and MuseNet \citep{payne2019musenet} to ABC-notation models with bar/patch-aligned hierarchical decoding \citep{wu2023tunesformer, qu2024mupt}. NotaGen \citep{wang2025notagen} scales this lineage with LLM-style pretraining and preference optimization, and is the backbone for our experiments. None of these models provide control over thematic material.

%\subsection*{Controllable symbolic generation.}
Control has been investigated for deep learning music models via control tokens \citep{keskar2019ctrl}, learned description sequences \citep{vonrutte2023figaro}, text conditioning \citep{lu2023musecoco}, and hard structural constraints \citep{herremans2017morpheus}. While interpretable probabilistic frameworks give users direct, musically meaningful handles \citet{hahn2023interpretable,hahn2024senthymnent, nihahn2025progress}, they do not leverage the musical abilities of foundational music LLMs. The closest work to ours is Theme Transformer \citep{shih2022theme}, which conditions on a concrete theme involving specific pitches and rhythms and is trained from scratch. We instead condition on an \emph{abstract, transposition-invariant} pattern and retrofit the capability onto a pretrained LLM while preserving its fluency.

\subsection*{Attention and Activation Steering}
Classifier-free guidance \citep{ho2022classifier}, post-hoc attention steering (PASTA) \citep{zhang2024pasta}, and activation addition \citep{turner2023activation} modify a \emph{frozen} model's computation. Our bias is mechanically similar to PASTA's logit offset but is applied during training as well as inference, so the model learns to exploit the emphasized tokens. Our two-phase schedule is a curriculum \citep{bengio2009curriculum} in which short focused crops make the motif-content association explicit and direct before full-length scores dilute training with additional musical material.

\subsection*{Motif Discovery in Music Information Retrieval (MIR)} 

Repeated-pattern discovery is a classical MIR problem, from geometric methods such as SIA/SIATEC \citep{meredith2002algorithms} and their compression-oriented successors \citep{collins2013siarct} to interval-based melodic pattern analysis \citep{conklin2001patterns} and closed-pattern motivic analysis \citep{lartillot2005efficient}. The task has also been studied directly on audio \citep{nieto2014identifying} and remains an active benchmark problem with dedicated datasets \citep{hsiao2023bps} (see \citet{janssen2013discovering} for a survey). Our contribution is not the discovery algorithm, which, while novel in its representation of the motif, is a simple sliding-window extractor, but rather its use as an effective control signal for generation.

\section{Methodology}
\label{sec:method}

\subsection{Motif Representation and Extraction}
\label{sec:representation}

\begin{figure}[t]
  \centering
  \includegraphics[width=0.95\linewidth]{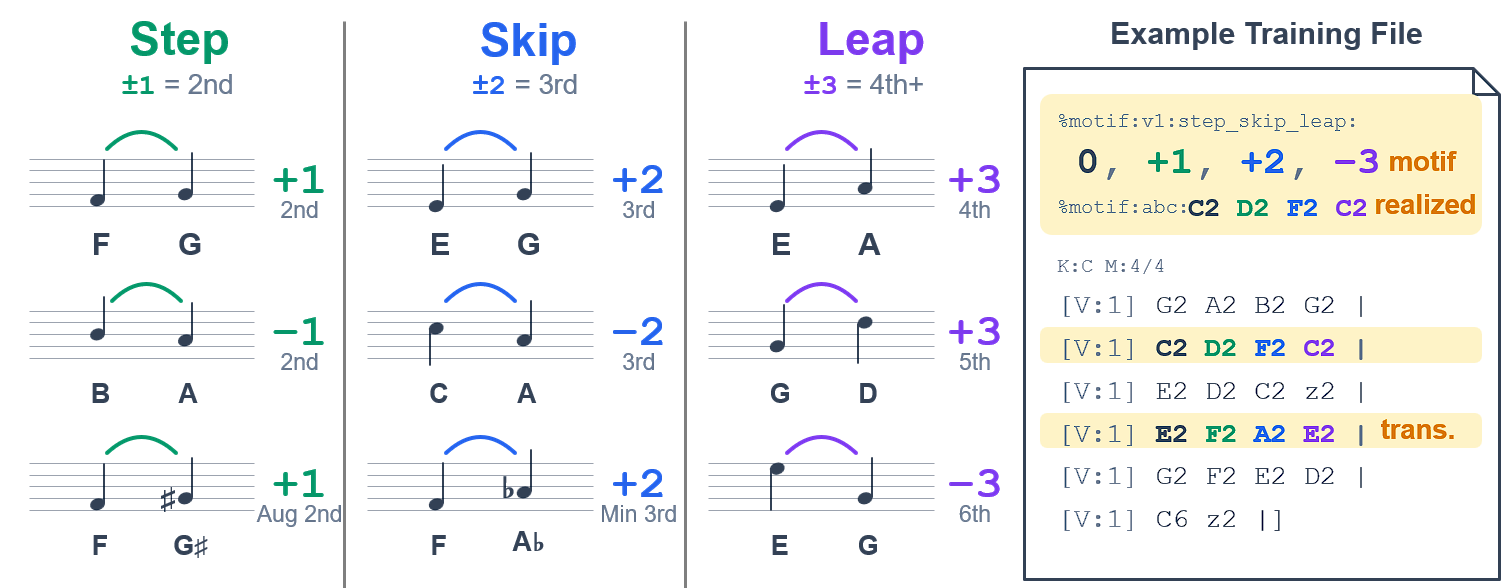}
  \caption{Interval-class encoding. Left: signed letter-step distances map to
  steps ($\pm1$), skips ($\pm2$), and leaps ($\pm3$). The classification is
  diatonic, so an augmented second is a step while its enharmonic minor third
  is a skip. Right: the resulting \texttt{\%motif:} prompt lines in a
  training file, with the realization and a transposed recurrence of the
  same pattern highlighted.}
  \label{fig:encoding}
\end{figure}

A motif must survive transposition and transformation. The opening four notes of Beethoven's Fifth is the same recognizable motif whether it starts on G or F, outlines a major or minor 3rd, or uses quarter notes instead of eighths. We therefore represent a motif not as pitches but as a sequence of \emph{signed interval classes} between consecutive melody notes, after collapsing immediate repetitions (see Figure~\ref{fig:encoding}). We measure the distance $d(\cdot,\cdot)\in\mathbb{Z}$ between two notes ($n_i$, $n_j$) in terms of diatonic scale steps. For example, $d(A\flat4,A\flat4)=0$, $d(A\flat3,B\flat3)=1$, and $d(C5,F4)=-4$. Intervals are then classified as \textit{steps} when $|d|=1$, \textit{skips} when $|d|=2$, and \textit{leaps} when $|d|\geq3$. We focus on pitch-based motifs and ignore note repetitions, so an interval class for $d=0$ is not required. Rather, we use 0 as a placeholder for the first note of a motif. For example, \motif{0,3,-1,-1} encodes a starting note, a leap up, and two steps down, respectively. The representation is transposition-invariant by construction and deliberately coarse. It captures contour and interval category, which is what most listeners track when they recognize a motif across keys and harmonizations.

%Each diatonic letter-step distance $d$ maps to $\operatorname{sign}(d)\cdot\{1,2,3\}$ for a step ($|d|{=}1$), skip ($|d|{=}2$), or leap ($|d|{\geq}3$), with a leading $0$ anchoring the first note.

To extract the motif from a piece, we use a sliding-window parser. We scan each piece end-to-end using a window length of 4--10 notes, and keep a count of each note sequence's occurrence throughout each piece. We select the most common sequence (weighted by length to mitigate inherent bias towards shorter motifs) to be a defining motif of the piece. For our experiments, we focus on motifs of four notes. Repeats and ornamentation are ignored throughout this process in order to avoid fragmenting motif counts. This extraction technique is used to annotate training data and evaluate generated results (details in Appendix~\ref{app:extraction}).

%Interestingly, even with the weighting, we still found that most pieces tend to produce motifs of length 4. Therefore, to reduce biases in data, avoid fragmentation, and limit burden on the model, the authors have deiced to use motifs of length 4 throughout the paper.

\subsection{Conditioning Interface}
\label{sec:conditioning}

\textbf{Novel Prompt Lines:} The motif is injected as structured comment lines prepended to the ABC header, replacing NotaGen's period--composer--instrumentation prompt. Concretely, we add \texttt{\%motif:v1:step\_skip\_leap: 0,3,-1,-1} as the abstract motif for voice \texttt{V:1}, and optionally \texttt{\%motif:abc: G B d c} as one concrete realization (see Figure~\ref{fig:encoding}, right). During inference, the model can be run \emph{abstract-only}, where the user provides the abstract motif, and the model will subsequent output the musical phrase containing the actual motif (the realization). Alternatively, the user could also provide the actual motif for \method{} to use.

\textbf{Motif Attention Bias:} A prompt line alone competes with hundreds of body tokens and its influence decays over a long generation. We therefore add a scalar bias $\beta$ to the pre-softmax attention logits at the motif prompt's key positions in the patch-level decoder $\alpha_{ij} = q_i^\top k_j/\sqrt{d} + \beta\cdot\mathbf{1}[j \in \mathcal{M}]$, where $\mathcal{M}$ is the set of patch positions belonging to \texttt{\%motif:} lines. The bias is applied identically at training and inference, so the model learns calibrated use of the emphasized channel rather than having its attention distorted post hoc (Figure~\ref{fig:training}, bottom). We find that the optimal bias strength for our case is $\beta{=}4$, while $\beta{=}2$ has little effect, and $\beta{\geq}8$ collapses generation entirely (more details in Section~\ref{sec:ablations}).

\subsection{Curriculum training}
\label{sec:curriculum}

\begin{figure}[t]
  \centering
  \includegraphics[width=0.95\linewidth]{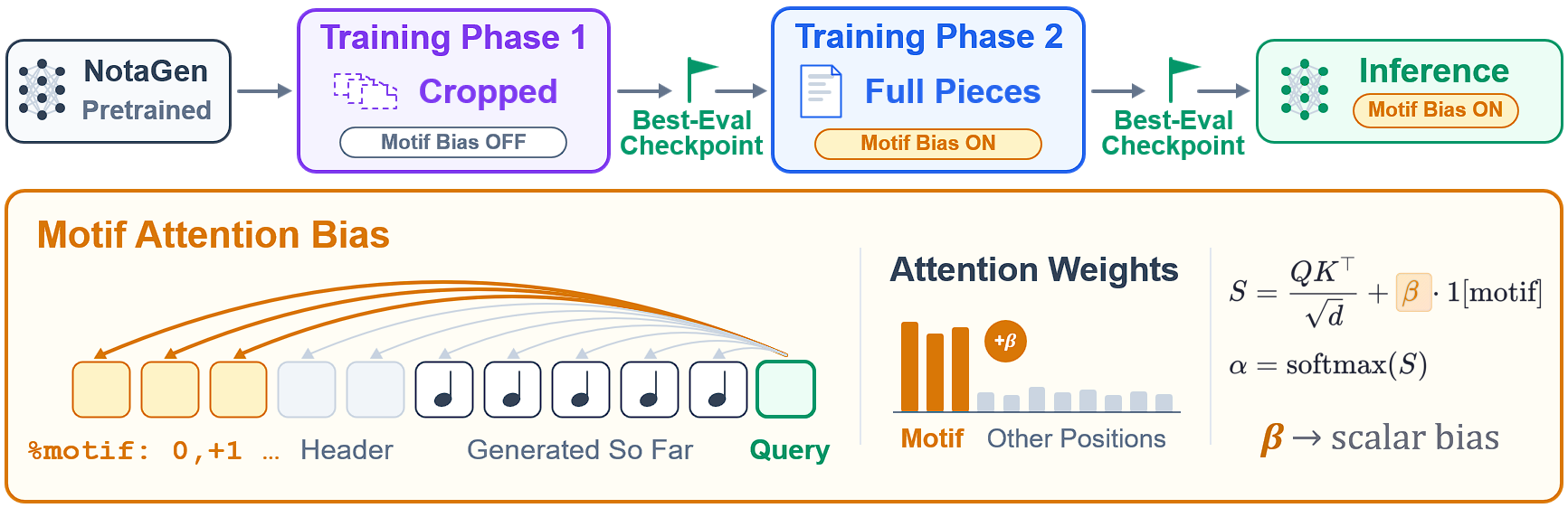}
  \caption{Two-phase curriculum with phase-matched evaluation (top) and an illustration of the
  motif attention bias (bottom), where a scalar $\beta$ added to the pre-softmax
  attention logits at the \texttt{\%motif:} key positions, identically at
  training and inference.}
  \label{fig:training}
\end{figure}

To ease the model into learning new prompt material, we structure training with a two-phase curriculum (overview in Figure \ref{fig:training}, top).

\textbf{Phase 1 (Focused Crops):} For every located motif occurrence in the source corpus we build focused cropped training data. The surrounding phrase, cropped to $\pm$5 measures around the occurrence, includes the full original header plus the motif prompt lines. Because these excerpts are real music centered on the motif instead of algorithmically generated samples, the distribution of training data does not drift far from full human-composed pieces. This phase allows the model learn the association between the motif prompts and the generated content of the music more directly.

\textbf{Phase 2 (Full Pieces):} We then fine-tune on full-length scores, each annotated with its extracted motif, with the attention bias active. Here, the model learns how the motif should be present and varied throughout the piece. Curriculum training implementation is detailed in Appendix \ref{app:curriculum_implementation}.

\subsection{Training Datasets and Motivic Analysis}
\label{sec:data}

We build on NotaGen \citep{wang2025notagen}, pretrained on $\sim$1.6M ABC-notation pieces and fine-tuned by its authors on $\sim$9k classical scores with period--composer--instrumentation prompts; all of our training modifies this checkpoint. Our source-data pool combines the OpenScore Lieder corpus \citep{gotham2022lieder} ($\sim$1.3k Romantic-era songs, from which we take the \texttt{V:1} vocal line) with the Irishman corpus \citep{irishman2023} of 216k+ monophonic folk tunes, each wrapped into a single-voice score so both sources share one format, these pieces are then augmented across 12 keys as shown in Figure~\ref{fig:overview}. The focused cropped phase-1 set consists of $\sim$650k+ motif-centered crops built from the same pool (Section~\ref{sec:curriculum}). Given the available dataset, the model will be trained to be monophonic only. In future works, when comparable or larger homophonic datasets becomes available, we plan to expand the model to generate multiple voices as well.
% TODO: pull final counts from build logs (corpus-composition table cut for now).
An exploratory analysis of the extracted motifs motivates our evaluation design: the source pool contains 210 distinct length-4 motif patterns whose frequencies are strongly skewed. Stepwise descent \motif{0,-1,-1,-1} occurs 62{,}356 times, the rank-27 pattern appears 1{,}083 times, and pure leap chains such as \motif{0,3,3,3} appearing twice in the entire pull (details in Appendix \ref{app:motif_analysis}). We therefore stratify our evaluation based motif frequency.
% Optional if space allows: small log-log rank--frequency figure (could go in
% the appendix instead).
%Any claim about motif conditioning therefore has to be stratified by training frequency, which our evaluation does explicitly.

\section{Experiments}
\label{sec:experiments}

%designed a series of evaluations and experiments to quantify the motif-related musical performance of the model. 

To evaluate \method{}, our experiments investigates how well \method{} can reliably compose music that organically integrates a prompted motif. We break this down into the following manageable questions: (1)~To what extent do our methods increase the model's performance? (Section~\ref{sec:ablations}), (2) how does performance scale with full-data quantity (Appendix~\ref{app:scaling}), and (3) Does \method{} incorporate the motif musically into the generated piece? We provide a case study of results in Section \ref{sec:case}, and plan for larger scale human evaluation in future works. 

% \subsection{Experimental setup}
% \label{sec:setup}

For each (model, motif) pair we generate 50 pieces with the abstract-only prompt and fixed decoding (top-$k{=}9$, top-$p{=}0.9$, temperature $1.2$), matching NotaGen's recommended sampling settings. The resulting samples are then evaluated using two metrics, one is \textit{body containment}, which asks whether the generated melody contains the target pattern anywhere, according to the motif extractor of Section~\ref{sec:representation}. Another is \textit{realization faithfulness}, which asks whether the model's own emitted \texttt{\%motif:abc} line realizes the target. Their $2\times2$ crosstab separates a model that understands a motif from one that further composes with it. The motifs used in experiments span the training distribution, namely we draw a stratified sample consisting of eight from the top-10 (band A) most frequent patterns, eight mid-distribution patterns (ranks 11--40 of 210, band B), and eight from ranks 41 and rarer (band C).

\subsection{Ablations}
\label{sec:ablations}

We ablate attention bias and curriculum training on the same protocol (Table~\ref{tab:ablation}). To test the efficacy of attention bias, we perform curriculum training with bias $\beta{=}0$ during both training and inference. To examine the benefits of curriculum training, we remove the cropped-data training phase, fine-tuning only on full pieces. Lastly, we judge performance after ablated both attention bias and cropped-data training phase. 

Results provide strong evidence that our full recipe is effective based on all three metrics. Furthermore, each ablated component is shown to positively contribute to motivic composition. Removing the attention bias costs the most while removing the cropped training phase has a milder effect on mid-distribution motifs. The cropped phase matters far more for motifs from the distribution tail and at small corpus scale (Appendix~\ref{app:scaling}). Interestingly, plain fine-tuning on the annotated corpus already reaches 79.4\% containment against a 35.9\% base rate. So it seems the prompt format itself is learnable from data alone, and the motif bias and curriculum training are complementary rather than additive; the cropped curriculum with the bias removed performs no better than plain fine-tuning, whereas together they lead on every metric.

% Latest experimental results
\begin{table}[t]
  \centering

  \small
  \begin{tabular}{lccc}
    \toprule
    Configuration & Containment & Faithfulness & Motifs / Measure \\
    \midrule
    \method{} (bias $\beta{=}4$ + curriculum) & \textbf{92.3\%} {\scriptsize[90.7, 93.7]} & \textbf{93.0\%} {\scriptsize[91.4, 94.3]} & \textbf{0.25} \\
    \quad $-$ attention bias                  & 77.8\% {\scriptsize[75.4, 80.1]} & 75.8\% {\scriptsize[73.3, 78.2]} & 0.18 \\
    \quad $-$ cropped phase                 & 86.7\% {\scriptsize[84.6, 88.5]} & 87.2\% {\scriptsize[85.2, 89.0]} & 0.22 \\
    \quad $-$ both                            & 79.4\% {\scriptsize[77.0, 81.6]} & 76.1\% {\scriptsize[73.6, 78.4]} & 0.19 \\
    Base rate (no motif prompt)               & 36.8\% {\scriptsize[34.1, 39.5]} & --- & 0.06 \\

    \bottomrule
  \end{tabular}
\caption{Ablation of \method{} components: body containment (the piece
  contains the motif at least once) and realization faithfulness, averaged over the 24 stratified sample motifs at $n{=}50$ per motif. $\beta{=}4$ configurations use training-matched bias application at inference. Bracketed values are $95\%$ Wilson score intervals on the pooled counts ($n{=}1200$). See Appendix ~\ref{sec:eval} for the definitions of Containment and Faithfulness. 
  \label{tab:ablation}}
\end{table}

\subsection{Case Study}
\label{sec:case}
\begin{figure}
    \centering
    \includegraphics[width=0.95\linewidth]{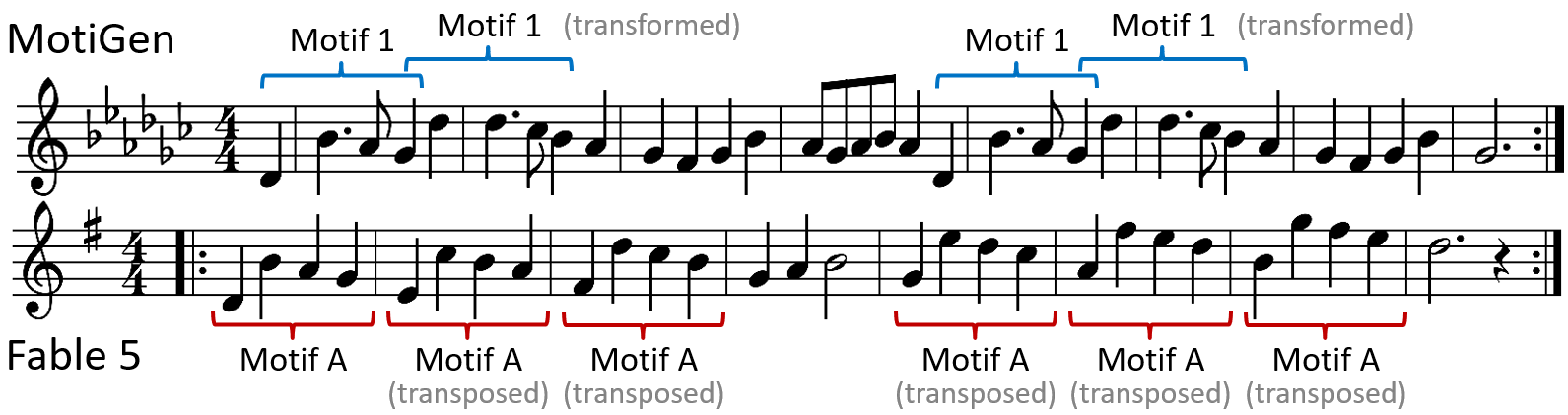}
    \caption{Excerpts from \method{} (upper) and Fable 5 (lower) when prompted to generate a melody with the motif of a leap up, and two steps down.}
    \label{fig:comparison}
\end{figure}

In order to get a better understanding of the subjective, qualitative qualities of the music generated by \method{}, both by itself, and in context of other methods, we provide a case study to examine a piece generated by \method{}. We prompt the model with a somewhat common, mid-distribution motif of (0, 3, -1, -1), which is a leap up, and two steps down, and allow the model to generate the realization, key signature, and time signature freely. 

Since no model to our knowledge generates music based on the same level of abstraction, we prompt a state-of-the-art general-purpose LLM, Fable 5 \citep{anthropic2026fable5}, with the following prompt and default settings to produce a comparable piece of music: 
\begin{verbatim}
        Generate a melody with a motif that goes a leap up
        and two steps down. Feel free to use any key signature
        or time signature, output in ABC notation.
\end{verbatim}
Figure \ref{fig:comparison} shows the melody generated by \method{} not only includes the motif as expected, but also includes interesting and diverse rhythms, as opposed to the simple quarter note sequence as created by the Fable 5 model. The realization of the motif is also more diverse in the sample generated by \method{}, as it includes both a leap with 6th (measure 1-2), and a 5th (measure 2), while the motifs in the sample by Fable 5 are all a 6th up and two steps down. Not to mention, the Fable 5 composition contains only the motif and nothing else other than the small cadence in measure 4, while \method{} has ample connecting materials between occurrences of the motifs.

\section{Conclusion}
\label{sec:conclusion}

We introduce \method{}, a recipe for retrofitting a thematic-control channel onto a pretrained symbolic music model by representing the motif as a transposition-invariant interval-class pattern, injecting it as a structured prompt line, reinforcing it with a matched train/inference attention bias, and training it with a focused-crop curriculum. Our resulting model composes convincing music with motivic material organically integrated into the texture rather than simply appended to the beginning. The same recipe could carry other musical instructions: harmonic progressions, bass lines, rhythmic cells, formal plans. %Every tool that has broadened who gets to make music started by shrinking the distance between an idea and the final product.

\bibliographystyle{plainnat}
\bibliography{references}

\appendix

\section{Implementation Details}
\label{app:curriculum_implementation}

Our code is available on Github\footnote{\url{https://github.com/cpyang123/motigen}}. For curriculum learning, checkpoints are selected by early stopping on held-out loss for up to 10 epochs, and evaluation is phase-matched. Each phase is evaluated on data from its own distribution and in its own bias configuration. Evaluating a bias-trained model with the bias off understates its quality and distorts the training curve; matching removes a spurious cross-phase jump that would otherwise suggest phase~2 begins by ``undoing'' phase~1. Figure~\ref{fig:training} summarizes the schedule.

All fine-tuning starts from the public NotaGen-large checkpoint (20-layer patch-level encoder, 6-layer character-level decoder, hidden size 1280, patch length 1024, patch size 16) in NotaGen's patch-streaming regime, using AdamW with learning rate $10^{-5}$ and batch size 1. Every training run fits on a single NVIDIA RTX A6000 (48\,GB); the reported experiments (the ablation grid of Table~\ref{tab:ablation}, the scaling grid of Appendix~\ref{app:scaling}, and their 50-piece-per-motif evaluation sweeps) total roughly under a GPU-month on this hardware.

\section{Data scaling improves motif adoption}
\label{app:scaling}

Beyond the main ablations, we ask whether motif adoption improves with more training data, holding the recipe fixed. To do this, we train all the model variants, ablated models and the full model, on two subsets of the training data. Specifically, a 10k subset that is an order of magnitude smaller than the 200k+ full dataset, and a 1k dataset that is another order of magnitude smaller still. Additionally, in order evaluate the models performance on motifs of different obscurity, we do the same stratified sample of 8 motifs from the top 10 motifs in the full dataset, middle 11-40 motifs, and the tail 41+ motifs, corresponding to A, B, and C respectively as in Table \ref{tab:scaling}. Each motif is sampled 50 times, and the score calculated as an average over all the generated pieces. The Full Training block reports the underlying component results of Table \ref{tab:ablation}.

\subsection{Full per-motif results}
\label{app:tables}

Closely examining Table \ref{tab:scaling} reveals quickly that the full advantage of the full training pipeline is greatest when we have few data points, second when we have relatively abundant data, and worse when the paucity of data greatly hinders generalization across all configurations. 

Utilizing the entirety of the training regiment almost unanimously provides better results when trained over the 10k subset and the full training dataset. This is especially prominent when looking at the generations using motifs from the tail-end of the distribution, those of rank C. When prompted with these motifs, the full training recipe produces almost consistently a 10\% improvement in both containment and faithfulness over the next best ablation configuration when trained on either the full and the 10k dataset. This could indicate a better internal understanding of the motivic structure and its relation to the generated piece, leading to better generalization on sparser data.

\begin{table}[h]
  \centering
  \small
  \begin{tabular}{lcccccc}
    \toprule
    & \multicolumn{3}{c}{Containment (\%)} & \multicolumn{3}{c}{Faithfulness (\%)} \\
    \cmidrule(lr){2-4}\cmidrule(lr){5-7}
    Configuration & A & B & C & A & B & C \\

    \midrule
    \multicolumn{7}{l}{\emph{$1$k source training pieces}} \\
    \method{} (full recipe)      & \textbf{55.0} {\tiny[50,\,60]} & 27.2 {\tiny[23,\,32]} & 19.0 {\tiny[15,\,23]} & \textbf{20.2} {\tiny[17,\,24]} & 4.2 {\tiny[3,\,7]} & 1.5 {\tiny[1,\,3]} \\
    \quad $-$ attention bias     & 54.0 {\tiny[49,\,59]} & 30.2 {\tiny[26,\,35]} & \textbf{20.5} {\tiny[17,\,25]} & 14.2 {\tiny[11,\,18]} & \textbf{4.8} {\tiny[3,\,7]} & \textbf{3.0} {\tiny[2,\,5]} \\
    \quad $-$ cropped phase      & 52.8 {\tiny[48,\,58]} & 29.0 {\tiny[25,\,34]} & 16.5 {\tiny[13,\,20]} & 15.0 {\tiny[12,\,19]} & 4.2 {\tiny[3,\,7]} & 1.0 {\tiny[0,\,3]} \\
    \quad $-$ both               & 53.5 {\tiny[49,\,58]} & \textbf{31.5} {\tiny[27,\,36]} & 16.0 {\tiny[13,\,20]} & 10.5 {\tiny[8,\,14]} & 3.2 {\tiny[2,\,5]} & 1.2 {\tiny[1,\,3]} \\
    \quad Baseline               & 47.0 {\tiny[42,\,52]} & 30.2 {\tiny[26,\,35]} & 14.5 {\tiny[11,\,18]} & --- & --- & --- \\
    \midrule
    \multicolumn{7}{l}{\emph{$10$k source training pieces}} \\
    \method{} (full recipe)      & \textbf{84.8} {\tiny[81,\,88]} & \textbf{64.5} {\tiny[60,\,69]} & \textbf{53.0} {\tiny[48,\,58]} & \textbf{84.5} {\tiny[81,\,88]} & \textbf{68.0} {\tiny[63,\,72]} & \textbf{59.2} {\tiny[54,\,64]} \\
    \quad $-$ attention bias     & 73.5 {\tiny[69,\,78]} & 61.0 {\tiny[56,\,66]} & 43.5 {\tiny[39,\,48]} & 58.2 {\tiny[53,\,63]} & 49.8 {\tiny[45,\,55]} & 38.0 {\tiny[33,\,43]} \\
    \quad $-$ cropped phase      & 68.5 {\tiny[64,\,73]} & 39.2 {\tiny[35,\,44]} & 21.2 {\tiny[18,\,26]} & 38.8 {\tiny[34,\,44]} & 21.0 {\tiny[17,\,25]} & 4.2 {\tiny[3,\,7]} \\
    \quad $-$ both               & 60.2 {\tiny[55,\,65]} & 46.8 {\tiny[42,\,52]} & 20.2 {\tiny[17,\,24]} & 21.8 {\tiny[18,\,26]} & 10.5 {\tiny[8,\,14]} & 3.2 {\tiny[2,\,5]} \\
    \quad Baseline               & 47.8 {\tiny[43,\,53]} & 32.5 {\tiny[28,\,37]} & 16.0 {\tiny[13,\,20]} & --- & --- & --- \\
    \midrule
    \multicolumn{7}{l}{\emph{Full Training}} \\
    \method{} (full recipe)      & \textbf{96.2} {\tiny[94,\,98]} & \textbf{94.0} {\tiny[91,\,96]} & \textbf{86.8} {\tiny[83,\,90]} & \textbf{95.2} {\tiny[93,\,97]} & \textbf{96.8} {\tiny[95,\,98]} & \textbf{87.0} {\tiny[83,\,90]} \\
    \quad $-$ attention bias     & 89.8 {\tiny[86,\,92]} & 86.5 {\tiny[83,\,90]} & 57.2 {\tiny[52,\,62]} & 85.5 {\tiny[82,\,89]} & 86.8 {\tiny[83,\,90]} & 55.2 {\tiny[50,\,60]} \\
    \quad $-$ cropped phase      & 95.8 {\tiny[93,\,97]} & 87.0 {\tiny[83,\,90]} & 77.2 {\tiny[73,\,81]} & 95.0 {\tiny[92,\,97]} & 87.2 {\tiny[84,\,90]} & 79.5 {\tiny[75,\,83]} \\
    \quad $-$ both               & 87.8 {\tiny[84,\,91]} & 84.0 {\tiny[80,\,87]} & 66.5 {\tiny[62,\,71]} & 82.2 {\tiny[78,\,86]} & 80.8 {\tiny[77,\,84]} & 65.2 {\tiny[60,\,70]} \\
    \quad Baseline               & 53.2 {\tiny[48,\,58]} & 38.0 {\tiny[33,\,43]} & 19.0 {\tiny[15,\,23]} & --- & --- & --- \\
    \bottomrule
  \end{tabular} 
  \caption{Training recipe crossed with source-data scale, on $24$ motifs
  stratified by corpus frequency (A: ranks 1--10; B: 11--40; C: 41--210;
  $n{=}50$ generations per motif, $8$ motifs per stratum, excluding erroneous generations. The baseline row is the unprompted Notagen model i.e.\ the rate at which the pattern appears by chance in its own idiom. Schedules are matched across scales; $\beta{=}4$ is used for attention bias. Bracketed values are $95\%$ Wilson score intervals on the pooled counts}
  \label{tab:scaling}
\end{table}

\section{Motif extraction details}
\label{app:extraction}

\subsection*{Pitch representation}
The key-signature-aware ABC parser tracks every note $n$ in two parallel coordinates: its \emph{chromatic} pitch $\pi(n) \in \mathbb{Z}$ (semitones, with the key signature and bar-persistent explicit accidentals applied) and its \emph{diatonic} staff position $\delta(n) = \ell(n) + 7\,o(n) \in \mathbb{Z}$, where $\ell(n) \in \{0,\dots,6\}$ is the letter index (C$\,{=}\,0$, D$\,{=}\,1$, \dots, B$\,{=}\,6$) and $o(n)$ the octave. $\delta$ depends only on the \emph{spelling} of the note, which is what lets the classifier distinguish enharmonically equivalent intervals.

\subsection*{Repeat collapse}
Given the note sequence $p_1,\dots,p_N$ of a voice, immediate repetitions are removed by keeping only the first note of each equal-pitch run: note $i$ is kept iff $i = 1$ or $\pi(p_i) \neq \pi(p_{i-1})$. A motif is a contour of \emph{distinct} successive notes, so a realization that restrikes a note inside the motif span (a tied or repeated tone) still counts as containing it.

\subsection*{Interval classification}
Write $q_1,\dots,q_n$ for the collapsed sequence, and for consecutive notes let $d_t = \delta(q_t) - \delta(q_{t-1})$ (signed letter-steps) and $s_t = \pi(q_t) - \pi(q_{t-1})$ (signed semitones). The interval class is
\begin{equation}
  c_t =
  \begin{cases}
    \operatorname{sign}(s_t) & \text{if } d_t = 0,\ s_t \neq 0,\\
    \operatorname{sign}(d_t)\cdot g(|d_t|) & \text{otherwise,}
  \end{cases}
  \qquad
  g(a) =
  \begin{cases}
    0 & a = 0,\\
    1 & a = 1 \text{ (step)},\\
    2 & a = 2 \text{ (skip)},\\
    3 & a \geq 3 \text{ (leap)},
  \end{cases}
  \label{eq:classify}
\end{equation}
where the first case counts a chromatic alteration of the same letter (e.g.\ F to F$\sharp$: $s_t \neq 0$ but $d_t = 0$) as a step, since the notes are distinct pitches; an interior $c_t = 0$ cannot occur because identical pitches are removed by the repeat collapse. Classification uses the letter-step distance $d_t$, so enharmonic spelling is honored: C to D$\sharp$ ($s_t{=}3$, $d_t{=}1$) is a step, while C to E$\flat$ ($s_t{=}3$, $d_t{=}2$) is a skip. A window of $k$ collapsed notes $(q_i,\dots,q_{i+k-1})$ maps to the pattern $m = (0, c_{i+1}, \dots, c_{i+k-1})$, the leading $0$ anchoring the first note. When spelling information is unavailable (plain semitone input, e.g.\ MIDI-derived), $d_t$ is approximated from $s_t$ by a nearest-interval table: $|s_t| \bmod 12 \in \{1,2\} \mapsto 1$, $\{3,4\} \mapsto 2$, $\{5\} \mapsto 3$, $\{6,7\} \mapsto 4$, $\{8,9\} \mapsto 5$, $\{10,11\} \mapsto 6$, plus $7$ per octave.

Since this method is interval-based, it is transposition-invariant. Transposing a passage adds a constant $(\Delta\delta, \Delta\pi)$ to every note's coordinates, leaving all differences $(d_t, s_t)$ --- and hence the pattern $m$ --- unchanged. Empirically, a sweep over 2{,}353 pieces $\times$ 12 key transpositions produced identical extracted motifs in all but 29 cases, each traced to inconsistencies in the source encodings rather than to the parser.

\subsection*{Extraction and containment}
Algorithm~\ref{alg:extract} gives the per-piece extraction. Containment checking at evaluation time reuses the same parse, collapse, and classification (Eq.~\ref{eq:classify}): a generated melody \emph{contains} target $m$ iff some length-$|m|$ window of its collapsed note sequence maps to $m$.

\begin{algorithm}[h]
  \caption{Per-piece motif extraction (window sweep 4--10, stride 1)}
  \label{alg:extract}
  \begin{algorithmic}[1]
    \Require ABC voice $V$; window range $w_{\min}{=}4$, $w_{\max}{=}10$
    \Ensure motif pattern $m^*$ and one concrete realization
    \State $(p_1,\dots,p_N) \gets \textsc{ParseABC}(V)$
      \Comment{per-note $(\pi, \delta)$ and bar index}
    \State $(q_1,\dots,q_n) \gets$ collapse immediate equal-pitch runs of $p$
    \For{$w = w_{\min}$ \textbf{to} $w_{\max}$}
      \For{$i = 1$ \textbf{to} $n - w + 1$}
        \State $m_i \gets (0, c_{i+1}, \dots, c_{i+w-1})$ by
          Eq.~\ref{eq:classify}
        \State $\mathrm{count}_w[m_i] \gets \mathrm{count}_w[m_i] + 1$;
          record occurrence (note span, bar)
      \EndFor
      \State $C_w \gets \max_m \mathrm{count}_w[m]$
    \EndFor
    \State $w^* \gets \arg\max_w C_w$ (smallest maximizing $w$);\quad
      $m^* \gets \arg\max_m \mathrm{count}_{w^*}[m]$
    \State \Return $m^*$ and the bar-aligned ABC snippet of its first
      occurrence
  \end{algorithmic}
\end{algorithm}

\subsection*{Evaluation metrics}
\label{sec:eval}
\textbf{Body containment} asks whether the model \emph{composed with} the requested motif. The generated melody is parsed, collapsed, and classified exactly as in training annotation (Eq.~\ref{eq:classify}), and the piece counts iff some length-$|m|$ window maps to the target pattern $m$. Any realization of the contour, in any key and anywhere in the piece, counts; the prompt lines are excluded from the search. We report the fraction of $n{=}50$ generations per motif that pass.

\textbf{Realization faithfulness} asks whether the model \emph{understood} the request. Given only the abstract pattern, the model opens its output with a \texttt{\%motif:abc:} line stating the notes it intends to use. Faithfulness parses that line alone with the same pipeline and checks that it maps to the target; a missing or malformed line counts as unfaithful. Since faithfulness reads only this header line and containment reads only the tune body, the two are independent, and their cross tabulation separates a model that merely echoes the instruction from one that uses it.

\section{Training Data Motivic Analysis}
\label{app:motif_analysis}

\begin{figure}[h]
  \centering
  \includegraphics[width=0.85\linewidth]{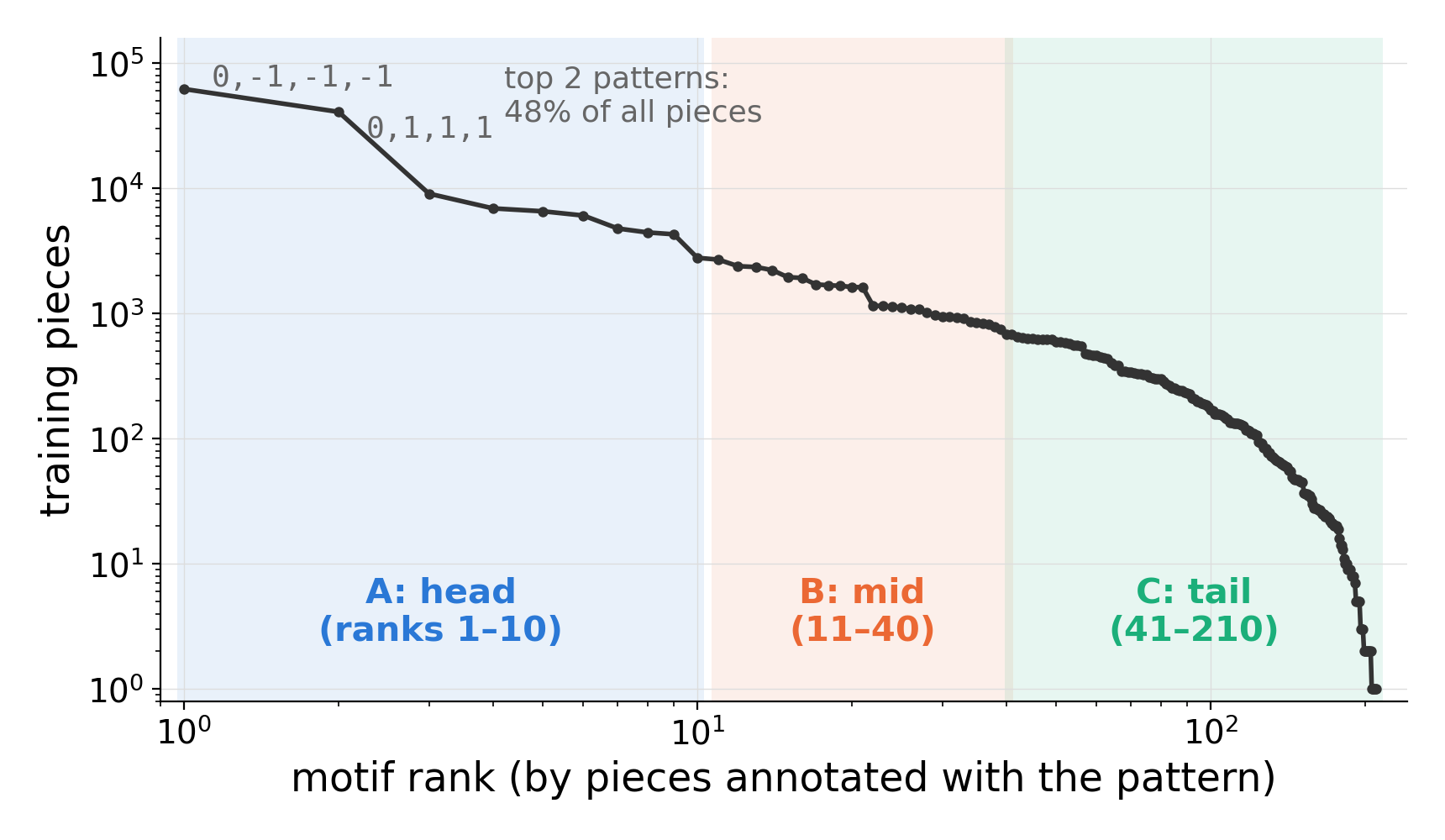}
  \caption{Motif rank frequency over the full training corpus (log log),
  ranked by how many pieces carry each pattern as their extracted motif
  annotation. Shaded bands mark the head, mid, and tail frequency regimes
  used for evaluation sampling.}
  \label{fig:zipf}
\end{figure}

Figure~\ref{fig:zipf} shows the rank frequency distribution of the extracted
motif annotations over the full training corpus (all Lieder plus the full
Irishman set). The distribution is heavily skewed: 210 distinct length-4
patterns occur across 217k annotated pieces, but the two stepwise runs
\motif{0,-1,-1,-1} and \motif{0,1,1,1} alone annotate 48\% of them, and
counts fall roughly geometrically through the remaining ranks. This skew is
what motivates the stratified evaluation of Appendix~\ref{app:scaling}:
results averaged over unstratified motifs would be dominated by patterns the
model has seen tens of thousands of times, hiding exactly the regime where
conditioning is hard. The shaded bands mark the head, mid, and tail
frequency regimes from which the evaluation motifs are drawn.

\section*{NeurIPS Paper Checklist}

\begin{enumerate}

\item {\bf Claims}
    \item[] Question: Do the main claims made in the abstract and introduction accurately reflect the paper's contributions and scope?
    \item[] Answer: \answerYes{}
    \item[] Justification: The abstract and introduction clearly state the main claims and contributions of the paper, accurately reflecting the theoretical and experimental results presented.

\item {\bf Limitations}
    \item[] Question: Does the paper discuss the limitations of the work performed by the authors?
    \item[] Answer: \answerYes{}
    \item[] Justification:  The paper mentions where limitations are met and how they are overcome throughout the paper.

\item {\bf Theory assumptions and proofs}
    \item[] Question: For each theoretical result, does the paper provide the full set of assumptions and a complete (and correct) proof?
    \item[] Answer: \answerNA{}
    \item[] Justification: The paper contains neither theoretical claims nor results.

    \item {\bf Experimental result reproducibility}
    \item[] Question: Does the paper fully disclose all the information needed to reproduce the main experimental results of the paper to the extent that it affects the main claims and/or conclusions of the paper (regardless of whether the code and data are provided or not)?
    \item[] Answer: \answerYes{}
    \item[] Justification: All information and artifacts needed to reproduce the results are provided.

\item {\bf Open access to data and code}
    \item[] Question: Does the paper provide open access to the data and code, with sufficient instructions to faithfully reproduce the main experimental results, as described in supplemental material?
    \item[] Answer: \answerYes{}
    \item[] Justification: Both the data and foundational model used are publicly known and available. Code is provided in Appendix A.

\item {\bf Experimental setting/details}
    \item[] Question: Does the paper specify all the training and test details (e.g., data splits, hyperparameters, how they were chosen, type of optimizer) necessary to understand the results?
    \item[] Answer: \answerYes{}
    \item[] Justification: All details are provided in Appendix A.

\item {\bf Experiment statistical significance}
    \item[] Question: Does the paper report error bars suitably and correctly defined or other appropriate information about the statistical significance of the experiments?
    \item[] Answer: \answerYes{}
    \item[] Justification: The experimental results provided include confidence intervals needed to claim statistical significance.

\item {\bf Experiments compute resources}
    \item[] Question: For each experiment, does the paper provide sufficient information on the computer resources (type of compute workers, memory, time of execution) needed to reproduce the experiments?
    \item[] Answer: \answerYes{}
    \item[] Justification: All resources needed are detailed in Appendix~\ref{app:curriculum_implementation}.
    
\item {\bf Code of ethics}
    \item[] Question: Does the research conducted in the paper conform, in every respect, with the NeurIPS Code of Ethics \url{https://neurips.cc/public/EthicsGuidelines}?
    \item[] Answer: \answerYes{}
    \item[] Justification: The research in the paper conforms with the NeurIPS Code of Ethics.

\item {\bf Broader impacts}
    \item[] Question: Does the paper discuss both potential positive societal impacts and negative societal impacts of the work performed?
    \item[] Answer: \answerYes{}
    \item[] Justification: The paper discusses and demonstrates the importance and viability of models built to take musical inputs intuitive to composers. 
    
\item {\bf Safeguards}
    \item[] Question: Does the paper describe safeguards that have been put in place for responsible release of data or models that have a high risk for misuse (e.g., pre-trained language models, image generators, or scraped datasets)?
    \item[] Answer: \answerNA{}
    \item[] Justification: The paper poses no such risks.

\item {\bf Licenses for existing assets}
    \item[] Question: Are the creators or original owners of assets (e.g., code, data, models), used in the paper, properly credited and are the license and terms of use explicitly mentioned and properly respected?
    \item[] Answer: \answerYes{}
    \item[] Justification: All data and models used are properly cited and credited. 

\item {\bf New assets}
    \item[] Question: Are new assets introduced in the paper well documented and is the documentation provided alongside the assets?
    \item[] Answer: \answerYes{}
    \item[] Justification: The code and demo website is provided in Appendix~\ref{app:curriculum_implementation} and the Abstract respectively.

\item {\bf Crowdsourcing and research with human subjects}
    \item[] Question: For crowdsourcing experiments and research with human subjects, does the paper include the full text of instructions given to participants and screenshots, if applicable, as well as details about compensation (if any)?
    \item[] Answer: \answerNA{}
    \item[] Justification: The paper involves no crowdsourcing or human-subject experiments. All evaluation is automatic, and a human listening study is explicitly deferred to future work (Section~\ref{sec:experiments}).

\item {\bf Institutional review board (IRB) approvals or equivalent for research with human subjects}
    \item[] Question: Does the paper describe potential risks incurred by study participants, whether such risks were disclosed to the subjects, and whether Institutional Review Board (IRB) approvals (or an equivalent approval/review based on the requirements of your country or institution) were obtained?
    \item[] Answer: \answerNA{}
    \item[] Justification: No research with human subjects was conducted.

\item {\bf Declaration of LLM usage}
    \item[] Question: Does the paper describe the usage of LLMs if it is an important, original, or non-standard component of the core methods in this research? Note that if the LLM is used only for writing, editing, or formatting purposes and does \emph{not} impact the core methodology, scientific rigor, or originality of the research, declaration is not required.
    %this research?
    \item[] Answer: \answerNA{}
    \item[] Justification: The paper does not involve LLMs as any important, original, or non-standard component.

\end{enumerate}

\end{document}